%% file: main.tex
\documentclass[10pt,twocolumn,letterpaper]{article}

\usepackage{3dv}
\usepackage{times}
\usepackage{epsfig}
\usepackage{graphicx}
\usepackage{amsmath}
\usepackage{amssymb}
\usepackage{mathalpha}
\usepackage{float}
\usepackage{cuted}
\usepackage[utf8]{inputenc}

\usepackage[pagebackref=true,breaklinks=true,colorlinks,bookmarks=false]{hyperref}

\usepackage[dvipsnames]{xcolor}
\usepackage[numbers,sort]{natbib}
\usepackage{tabularx}
\usepackage{mathtools}
\usepackage{booktabs}
\usepackage{nonfloat}
\usepackage{enumitem}
\usepackage{verbatim}
\usepackage{subcaption}
\usepackage{amsfonts} 
\DeclareMathSymbol{\shortminus}{\mathbin}{AMSa}{"39}
\DeclareFontFamily{OT1}{pzc}{}
\DeclareFontShape{OT1}{pzc}{m}{it}{<-> s * [1.150] pzcmi8t}{}
\DeclareMathAlphabet{\mathpzc}{OT1}{pzc}{m}{it}
\usepackage[mathscr]{eucal}
\usepackage{comment}
\usepackage{soul}
\usepackage{xfakebold}

\newcommand{\fbseries}{\unskip\setBold\aftergroup\unsetBold\aftergroup\ignorespaces}
\makeatletter
\newcommand{\setBoldness}[1]{\def\fake@bold{#1}}
\makeatother

\definecolor{cadmiumgreen}{rgb}{0.0, 0.42, 0.24}

\newcommand{\francesc}[1]{{\color{orange}#1}}

\DeclareMathSymbol{\shortminus}{\mathbin}{AMSa}{"39}
\newcommand{\fig}[1]{Fig.~\ref{fig:#1}}
\newcommand{\sect}[1]{Sect~\ref{sect:#1}}
\newcommand{\chapt}[1]{Chapter~\ref{chapter:#1}}
\newcommand{\tab}[1]{Table~\ref{tab:#1}}
\newcommand{\alg}[1]{Algorithm~\ref{alg:#1}}
\newcommand{\eq}[1]{Eq. (\ref{eq:#1})}
\newcommand{\app}[1]{Appendix~\ref{app:#1}}
\newcommand\RGB{\textcolor{black}{R}\textcolor{black}{G}\textcolor{black}{B }}

\newcommand\srx[1]{\textbf{\textcolor{red}{ShahRukh: #1}}}
\newcommand\tentative[1]{\textit{\textcolor{red}{tentative: #1}}}

\newcommand\albert[1]{\textbf{\textcolor{blue}{Aggelina: #1}}}
\newcommand\ds[1]{\textbf{\textcolor{orange}{Dimitris: #1}}}
\newcommand\textttmw[1]{\texttt{{\fbseries #1}}}

\usepackage{xspace}

\usepackage[pagebackref=true,breaklinks=true,colorlinks,bookmarks=false]{hyperref}
\newcommand{\MethodName}{SIDER\xspace}

\newcommand{\Ibf}{\mathbf{I}}
\newcommand{\xbf}{\mathbf{x}}
\newcommand{\pixbf}{\mathbf{p}}
\newcommand{\alphabf}{\boldsymbol\alpha}
\newcommand{\gammabf}{\pmb{\gamma}}
\newcommand{\epbf}{\pmb{\epsilon}}

\def\mbf#1{\mathbf{#1}}
\def\mbb#1{\mathbb{#1}}
\def\mcal#1{\mathcal{#1}}
\def\mtxt#1{\text{#1}}
\def\mtxtlog{\text{log }}
\usepackage{environ}
\NewEnviron{smequation}{
    \normalsize
    \begin{equation}
        \BODY
    \end{equation}
}
\NewEnviron{smequation*}{%
    \begin{equation*}
    \scalebox{0.95}{$\BODY$}
    \end{equation*}
}
\makeatletter
\newcommand{\printfnsymbol}[1]{%
  \textsuperscript{\@fnsymbol{#1}}%
}
\makeatother

\threedvfinalcopy 

\ifthreedvfinal\fi
\DeclareUnicodeCharacter{1EA7}{\(\hat{a}\)}
\DeclareUnicodeCharacter{1EA5}{\(\hat{a}'\)}
\begin{document}
\newcommand*\samethanks[1][\value{footnote}]{\footnotemark[#1]}
\title{\MethodName \includegraphics[height=15pt]{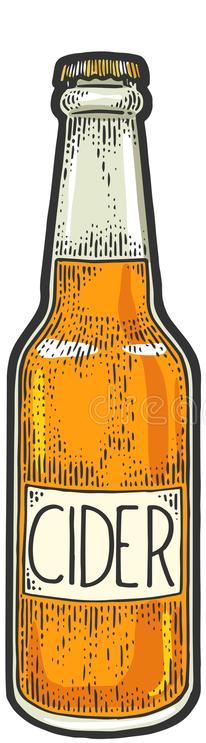}: Single-Image Neural Optimization for Facial Geometric Detail Recovery}

\author{
Aggelina Chatziagapi\(^{1}\)\thanks{Equal Contribution }
\and
ShahRukh Athar\(^{1}\)\samethanks
\and 
Francesc Moreno-Noguer$^{2}$
\and
Dimitris Samaras$^{1}$
\\~\\
${ }^{1}$Stony Brook University\\
${ }^{2}$Institut de Rob\`{o}tica i Inform\`{a}tica Industrial, CSIC-UPC
}

\maketitle

\begin{abstract}
   We present \MethodName~(\textbf{S}ingle-\textbf{I}mage neural optimization for facial geometric \textbf{DE}tail \textbf{R}ecovery), a novel photometric optimization method that recovers detailed facial geometry from a single image in an unsupervised manner. Inspired by classical techniques of coarse-to-fine optimization and recent advances in implicit neural representations of 3D shape, \MethodName combines a geometry prior based on  statistical models and Signed Distance Functions (SDFs) to recover facial details from single images. First, it estimates a coarse geometry using a morphable model represented as an SDF. Next, it reconstructs facial geometry details by optimizing a photometric loss with respect to the ground truth image. In contrast to prior work, \MethodName does not rely on any dataset priors and does not require  additional supervision from multiple views, lighting changes or ground truth 3D shape. Extensive qualitative and quantitative evaluation demonstrates that our method achieves state-of-the-art on facial geometric detail recovery, using only a single in-the-wild image. 
\end{abstract}

\section{Introduction}
The study of the 3D geometry of the human face is a problem of great interest in computer graphics and vision communities. Early approaches to recover the 3D structure of the human face were based on morphable models \cite{blanz1999morphable}, where the face shape, expression and texture are optimized with respect to a given image. However, due to the low dimensionality of the identity and expression subspaces, morphable models are unable to capture facial geometric details, such as wrinkles and skin-folds. With the advent of deep learning, it has now become possible to train networks on large datasets \cite{DECA, tewari2019fml, sela2017unrestricted}, in order to regress the 3D shape of the face along with its geometric details. Still, the generalization ability of these methods is limited, mainly due to the lack of diversity of the data they are trained on~\cite{KANface}.

Recent advances in implicit neural representations of 3D shape \cite{IDR, siren, IGR} have made it possible to learn rich details of the 3D geometry of an object. The geometry can be represented by Signed Distance Functions (SDFs) \cite{IDR, siren, IGR}, Unsigned Distance Functions (USDFs) \cite{USDFs} or occupancies \cite{OccNet} that are parameterized using Multilayer Perceptrons (MLPs). The high representational power of MLPs, along with the use of positional encoding \cite{nerf}, facilitates the reconstruction of rich geometric details. Additionally, if the geometry is represented as an SDF, recent works \cite{DVR, IDR} have proposed ways that allow the cheap calculation of derivatives of the geometry through implicit differentiation, making gradient learning significantly more tractable. These implicit representations are learnt using either multi-view supervision \cite{IDR, DVR} or partial 3D data \cite{siren, IGR}. However, in the absence of 3D data or multi-view supervision, like when one has access only to a single face image, it might not be possible to train such models as they may collapse into trivial solutions.

In this work, we propose \MethodName, a novel photometric optimization method that recovers facial geometric details from a single face image. \MethodName uses an unsupervised coarse-to-fine optimization scheme that does not require any ground truth 3D, multi-view or varying light-source supervision. Since optimization of SDFs using a single image is prone to trivial solutions, SIDER first learns a coarse approximation of the 3D face geometry using a morphable model fit to the input image by a standard landmark fitting pipeline. Next, this SDF is optimized by minimizing the photometric loss with respect to the given image via implicit differentiation \cite{IDR, DVR}. After converging, \MethodName outputs an SDF that represents the 3D face shape along with its geometric details (see Fig 1). We show, both quantitatively and qualitatively, that \MethodName significantly outperforms the current state-of-the-art in detailed face reconstruction by recovering facial geometric details that are realistic and have high fidelity to the input image.  


To summarize, our contributions are as follows:
\begin{itemize}
    \item{We propose \MethodName, a method that recovers facial geometric details from a single face image in an unsupervised manner}.
    \item{We propose a novel coarse-to-fine optimization scheme that leverages a classical morphable model representation as a prior to prevent degenerate solutions of the SDF and is optimized using an unsupervised photometric loss.}
    \item{We achieve state-of-the-art performance in facial geometry reconstruction from single in-the-wild images.}
\end{itemize}

\input{related}
\input{MethodsAndExp}

\section{Conclusion}
In this work we present \MethodName, a method for high-fidelity detailed 3D face reconstruction from a single image that can be trained in an unsupervised manner. Our approach combines  the best from classical statistical models and recent implicit neural representations. The former is used to obtain a coarse shape prior, and the latter provides high frequency detail of the geometry, by only optimizing over a photometric loss computed w.r.t. the input image. A thorough quantitative and qualitative evaluation, shows that \MethodName  outperforms current state-of-the-art by a significant margin. A limitation of our current approach is that it still cannot handle details like hair or beards. This is because the photometric loss for these regions would require sub-pixel accuracy. In the future, we will explore alternatives for addressing this type of high-frequency details.

{\small
\bibliographystyle{ieee_fullname}
\bibliography{egbib}
}

\onecolumn
\input{suppmat_arxiv}

\end{document}

%% file: related.tex
\section{Related Work}
In this section, we describe recent related works in facial geometry estimation, facial geometric details recovery and implicit representations of 3D shapes.

\begin{figure*}[ht!]
    \centering
    \includegraphics[width=1.0\linewidth]{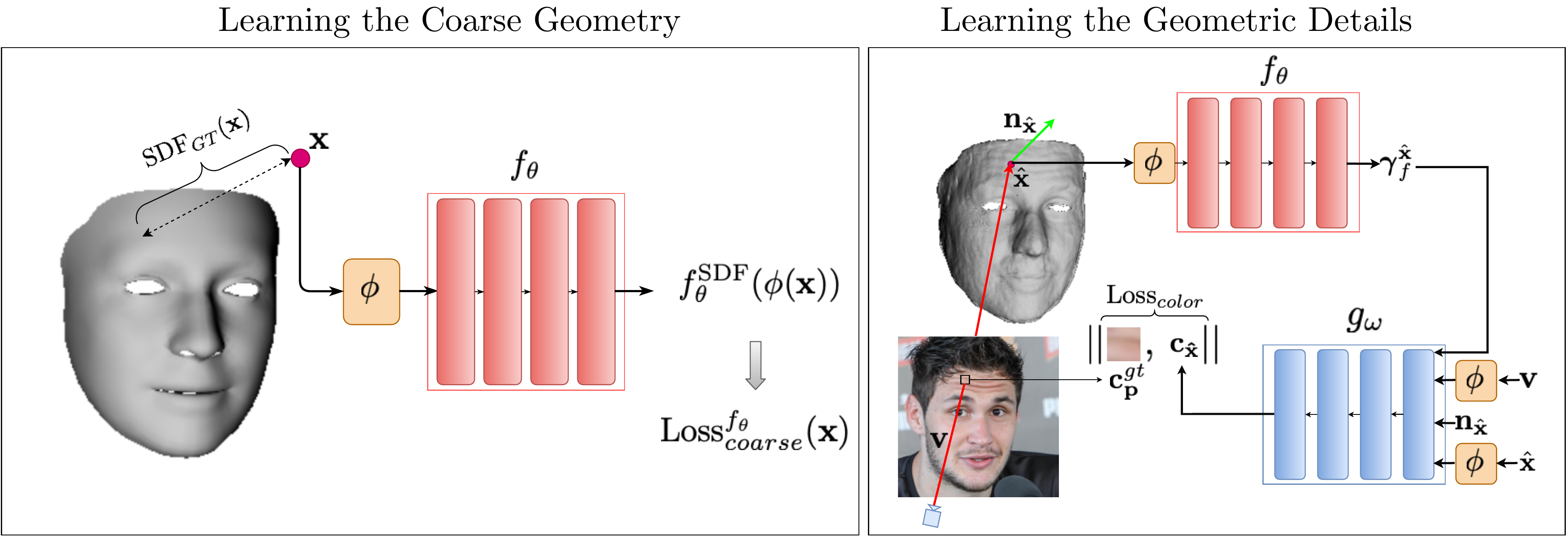}
    \caption{\small{\textbf{Overview of \MethodName:} {\bf Left:} First \MethodName learns a coarse geometry, represented as an SDF, by using FLAME mesh \cite{FLAME:SiggraphAsia2017} fit to the input image as supervision {\bf Right:} Next, it recovers facial geometric details by optimizing the SDF w.r.t. the photometric loss on the input image.}} 
    \label{fig:architecture}
\end{figure*}

\vspace{1mm}
\noindent{\bf Facial Geometry Estimation.} One of the first widely used methods for 3D shape reconstruction of the human face was based on statistical 3D face models that can fit a given image, going back to the original 3D Face Morphable Model~\cite{blanz1999morphable} (3DMMs). However, the underlying PCA-based representation for shape and expression of 3DMMs is not flexible enough to represent fine facial geometric details such as wrinkles, skin folds and skin bumps. Recent methods \cite{tewari2019fml, tewari2017self, tran2018nonlinear, tran2019towards, reda, booth20173d, dou2017end, jackson2017large, kim2018inversefacenet, Genova_2018_CVPR} leverage the power of deep-learning and large-scale image and video datasets  to regress parameters that generate realistic 3D reconstructions or learn  complex representations  for face shape, expression and texture. These methods produce more realistic results than traditional 3DMM fitting, but they still  cannot capture fine facial details and need large datasets to train on. In contrast, we propose a neural coarse-to-fine optimization scheme to recover facial geometric details from single images and no-supervision.

\vspace{1mm}
\noindent{\bf Geometric Facial Details Estimation.}  The past few years have witnessed a significant improvement in the realism of reconstructed 3D face geometry and the quality of facial details. In \cite{Richardson2017FaceRF}, a CNN (CoarseNet) first regresses a rough geometry of the face, and then another CNN (FineNet) estimates facial details using a coarse depth map and input images. In \cite{sela2017unrestricted}, the  regressed correspondence and depth maps are registered onto a template mesh, which   is further refined to generate the detailed facial geometry. In \cite{extreme3D}   facial details are modelled with bump maps on top of a 3DMM base. DF$^{2}$Net~\cite{df2net} uses multiple refinement steps to reconstruct detailed geometric structure. First, a coarse depth is predicted, which is then refined by an F-Net. The refined depth along with the input image is then given as input to a specially designed Finer-Net that outputs the final recovered facial details. DECA~\cite{DECA} uses a differentiable renderer to perform 3D face reconstruction and recover the detailed face geometry. The facial geometric details are represented as a UV-map of vertex displacements of a FLAME mesh\cite{FLAME:SiggraphAsia2017}. DECA is trained on a large dataset of close to 2 million images with a subset of them being paired. In contrast to the aforementioned methods, \MethodName does not require an exorbitantly large dataset for training and can be used on single in-the-wild images.

\vspace{1mm}
\noindent{\bf Implicit Neural Representations.} Representing 3D shapes implicitly with neural networks, more specifically using Multi-Layer Perceptrons (MLPs), has led to the development of methods that are able to reconstruct a large variety of 3D shape with rich details \cite{IDR, siren, IGR, OccNet, DVR, Liu2019LearningTI, nglod, DeepSDF, USDFs}. In \cite{DeepSDF}, authors train an SDF using instance specific meshes. Once trained, a latent space for the instances is learnt, making it possible to sample a large variety of shapes represented as SDFs. In \cite{IGR}, the authors propose a regularizer, the eikonal constraint, that allows learning SDFs from sparse 3D samples in the form of a point cloud. The authors of \cite{IDR} use the eikonal constraint proposed by \cite{IGR} to learn 3D shapes of objects using multi-view supervision along with object masks. They also propose an expression for the derivative of the intersection point of sphere tracing with respect to the MLP parameters that matches the real derivative up to the first order. \MethodName uses the expression for the derivative of the intersection point proposed by \cite{IDR}, in order to back-propagate gradients to its geometry network and recover the facial geometric details. However,  unlike the aforementioned methods, \MethodName leverages on a prior estimated from statistical models, which constrains the optimization and allows recovery of facial geometric details from single images without any multi-view supervision or ground truth 3D data.

%% file: MethodsAndExp.tex
\section{\MethodName}

\begin{figure*}[h!]
\centering
\includegraphics[width=\linewidth]{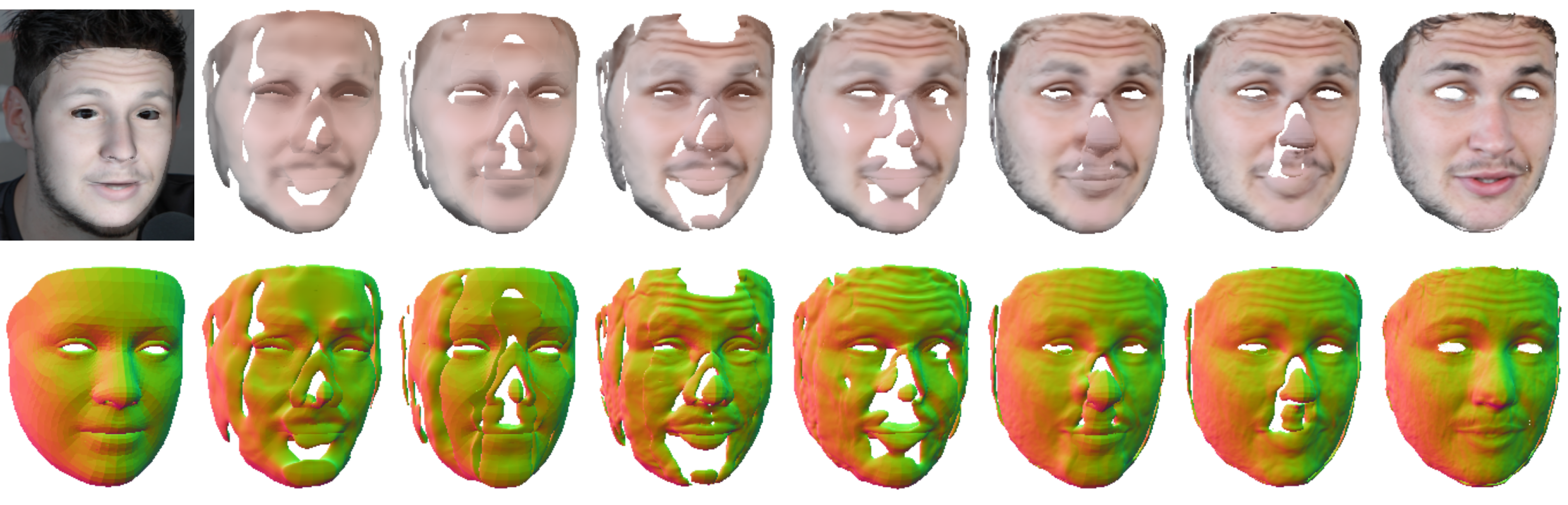}
\caption{\textbf{\MethodName in action:} Here we show the intermediate output of \MethodName as it is trained. The first column contains an overlay of the initial FLAME \cite{FLAME:SiggraphAsia2017} geometry on top of the image (top row) and the normals of the initial geometry. The subsequent columns show the rendered RGB (top row) and the normals of the learnt geometry (bottom row) as \MethodName is trained. Once converged, we get a facial geometry with high quality details (last column, bottom row) and its photrealistic rendering (last column, top row).}
\label{fig:sider_train}
\end{figure*}

Given a single image, \MethodName aims to extract from it facial geometric details, such as wrinkles and skin folds. \MethodName uses a two-stage neural optimization approach to extract these details. In the first stage, we leverage the FLAME morphable model  \cite{FLAME:SiggraphAsia2017} prior, in order to learn the coarse geometry of the face, which is represented as an SDF. Next, we optimize this SDF w.r.t the photometric loss of the provided image, \(\Ibf\), in order to learn the facial geometric details. In the following, we elaborate the workings and the training process of \MethodName.

\subsection{Architecture}
As shown in the overview Fig.~\ref{fig:architecture}, \MethodName consists of two MLPs: a geometry network \(f_{\theta}\) and a rendering network \(g_{\omega}\). The geometry MLP, \(f_{\theta}(\cdot)\) represents the face shape (along with the facial geometric details) as an SDF. More specifically, for any point \(\xbf\):
\begin{equation}
    f_{\theta}(\xbf) = \{f_{\theta}^{\text{SDF}}(\phi(\xbf)), \gammabf_{f}^{\xbf}\}
\end{equation}
\noindent where \(f_{\theta}^{\text{SDF}}(\xbf)\) is the SDF predicted at \(\xbf\), \(\gammabf_{f}^{\xbf}\) is a feature vector predicted at \(\xbf\) that is used as input to the rendering network \(g_{\omega}\), and \(\phi\) is the positional encoding.

The rendering network, \(g_{\omega}(\cdot)\), predicts the RGB value of a point \(\xbf\) as follows:
\begin{equation}
    g_{\omega}(\phi(\xbf), \mathbf{n}_{\xbf}, \phi(\mathbf{v}), \gammabf_{f}^{\xbf}) = \{R,G,B\}
\end{equation}
\noindent where \(\xbf\) are the point's coordinates, \(\mathbf{n}_{\xbf}\) is the normal at point \(\xbf\), \(\mathbf{v}\) is the viewing direction and \(\gammabf_{f}^{\xbf} =  f_{\theta}(\xbf)\) is a feature vector predicted by the geometry network \(f_{\theta}\) at \(\xbf\).

\subsection{Learning the coarse geometry}
\label{sect:coarse_learn}
Given a face image \(\Ibf \in \mbb{R}^{H \times W \times 3}\), \MethodName first learns a coarse geometry of the face using a morphable model prior. FLAME is fit to the face in \(\Ibf\) using standard landmark fitting \cite{DECA, 3DDFA_V2}
\begin{equation}
    \underset{\alphabf_{\text{shape}}, \alphabf_{\text{exp}}, \alphabf_{\text{pose}}, \textbf{cam}}{\text{min}} ||L_{FLAME}^{i} - L_{gt}^{i}||; \quad \forall i
\end{equation}

\noindent where \(L_{FLAME}^{i}\) is the position of the \(i\)'th landmark of the FLAME model \cite{FLAME:SiggraphAsia2017}, \(L_{gt}^{i}\) is the position of the \(i\)'th landmark predicted by 3DDFA \cite{3DDFA_V2} (which we treat as ground truth),  \(\alphabf_{\text{shape}}, \alphabf_{\text{exp}}, \alphabf_{\text{pose}}\) are the shape, expression and pose parameters of the FLAME model \cite{FLAME:SiggraphAsia2017} and \(\textbf{cam}\) are the camera parameters.
Once FLAME \cite{FLAME:SiggraphAsia2017} is fit to the image, we train an MLP, \(f_{\theta}\), to represent this coarse mesh as an SDF by minimizing the following:
\begin{equation}
    \text{Loss}_{geo}(\xbf) =  ||f_{\theta}^{\text{SDF}}(\phi(\xbf)) - \text{SDF}_{GT}(\xbf)||; \quad \forall \xbf \in \mathcal{P}
\end{equation}
\noindent where \(\mathcal{P}\) is a set of randomly chosen points in space in the neighborhood of the FLAME mesh,  \(\xbf = \{x,y,z\}\) is a point in space, \(\phi\) is the positional encoding and \(\text{SDF}_{GT}(\cdot)\) is the ground truth SDF to the coarse mesh. Since the FLAME face mesh is an open and single surface layer, 
an SDF cannot be defined on it directly (there is no region where the distance is negative, since there is no `inside'). Therefore, in order to define the SDF, we consider the face mesh to be volume of `thickness' \(\epbf\). This allows us to define the ground-truth SDF as follows
\begin{equation}
    \text{SDF}_{GT}(\xbf) = \text{Point2Mesh}(\xbf) - \frac{\epbf}{2}
\end{equation}
\noindent where \(\text{Point2Mesh}\) is the point-to-mesh distance function and \(\epbf\) is a small number denoting the thickness of the mesh. Additionally, the geometry network is regularized using the eikonal loss:
\begin{equation}
    \text{Loss}_{eik}(\xbf) = \mathbb{E}_{\xbf}(||\nabla_{\xbf}f_{\theta}(\xbf)|| - 1)^{2}
    \label{eq:eik}
\end{equation}
The full loss on the geometry network is
\begin{equation}
    \text{Loss}^{f_{\theta}}_{coarse}(\xbf) = \text{Loss}_{geo}(\xbf) + \lambda \text{Loss}_{eik}(\xbf)
    \label{eq:coarse}
\end{equation}
where \(\lambda\) is a regularization coefficient.

\begin{figure*}[h]
\begin{center}
\includegraphics[width=0.9\linewidth]{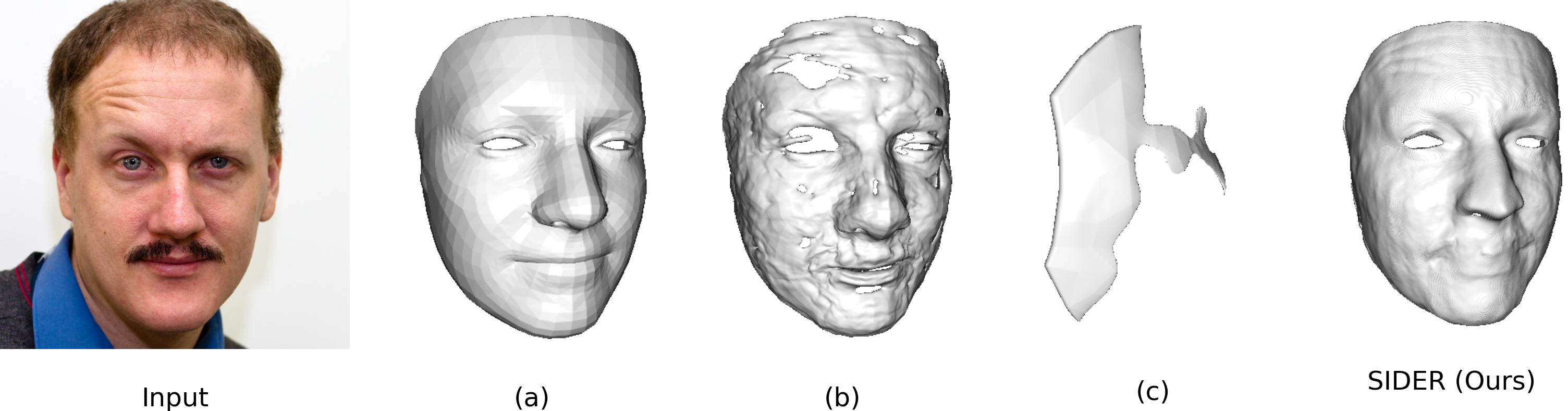}
\end{center}
    \caption{\textbf{Ablation Study}: Here we ablate the contributions of the normals and the feature vector as input to the gradient network (see \eq{color_loss}). The first column is the image on which \MethodName is optimized. (a) Geometry learnt without using the feature vector as input. (b) Geometry learnt without using the normals as input, (c) Geometry learnt without using both feature vector and normals as input. The last column shows the result of \MethodName that uses both the normals and the feature vector as input to the rendering network.}
\label{fig:ablation}
\end{figure*}

\subsection{Recovering the Facial Geometric Details}
Once \(f_{\theta}\) has been trained to approximate the coarse geometry, we use the provided image, \(\Ibf\), to fine-tune \(f_{\theta}\) and recover the facial geometric details. In order to render the SDF \(f_{\theta}\), we use sphere-tracing along with a rendering network \(g_{\omega}\). Rays are shot into the scene from the camera center \(\mathbf{o}\), and the intersection of these rays with the face mesh is estimated using sphere-tracing and implicit differentiation \cite{IDR}. The colors of the intersecting points are predicted using \(g_{\omega}\). More specifically, consider a ray, \(\mathbf{r} = \mathbf{o} + \mathbf{v}t\), with viewing direction \(\mathbf{v}\) and surface intersection point \(\hat{\xbf}\). The RGB color at \(\hat{\xbf}\) is calculated as follows:
\begin{equation}
    \mathbf{c_{\hat{\xbf}}} = g_{\omega}(\phi(\hat{\xbf}), \mathbf{n}_{\hat{\xbf}}, \phi(\mathbf{v}), \gammabf_{f}^{\hat{\xbf}}) 
\end{equation}
\noindent where \(\mathbf{c_{\hat{\xbf}}}\) is the predicted RGB color at point \(\hat{\xbf}\), \(\mathbf{n}_{\hat{\xbf}}\) is the normal at point \(\hat{\xbf}\) and \(\gammabf_{f}^{\hat{\xbf}}\) is a feature vector predicted by the geometry network, \(f_{\theta}\) at \(\hat{\xbf}\). The facial geometric details are recovered by jointly optimizing the geometry network \(f_{\theta}\) and the rendering network \(g_{\omega}\) with respect to the photometric loss as follows:
\begin{equation}
    \underset{\theta, \omega}{\text{min  }}\text{Loss}_{color}(\mathbf{c_{\hat{\xbf}}}, \mathbf{c}_{\pixbf}^{gt}) =  ||\mathbf{c_{\hat{\xbf}}} - \mathbf{c}_{\pixbf}^{gt}||
    \label{eq:color_loss}
\end{equation}
\noindent where \(\mathbf{c}_{\pixbf}^{gt}\) is the ground truth pixel color. The gradients to the geometry network \(f_{\theta}\), are calculated using implicit differentiation \cite{IDR}. Additionally, in order to ensure that \(f_{\theta}\) does not drift too far away from the face shape, it is regularized using the coarse SDF from \sect{coarse_learn}. The complete loss of the geometry network is:
\begin{equation}
    \begin{split}
         \text{Loss}^{f_{\theta}}_{detail}(\mathbf{c_{\hat{\xbf}}}, \mathbf{c}_{\pixbf}^{gt}) =  & ||\mathbf{c_{\hat{\xbf}}} - \mathbf{c}_{\pixbf}^{gt}|| \\
                            &+ \lambda_{1}  ||f_{\theta}(\phi(\xbf)) - \text{SDF}_{GT}(\xbf)|| \\
                            & + \lambda_{2} \text{Loss}_{eik}
    \end{split}
\end{equation}
where \(\lambda_{1}, \lambda_{2}\) are the regularization coefficients.

\section{Experiments}

In this section, we evaluate \MethodName's ability to recover facial geometric details from single in-the-wild images. We present both qualitative and quantitative results, and compare our proposed approach to the current state of the art.

\begin{figure*}[t]
\begin{center}
\includegraphics[width=\linewidth]{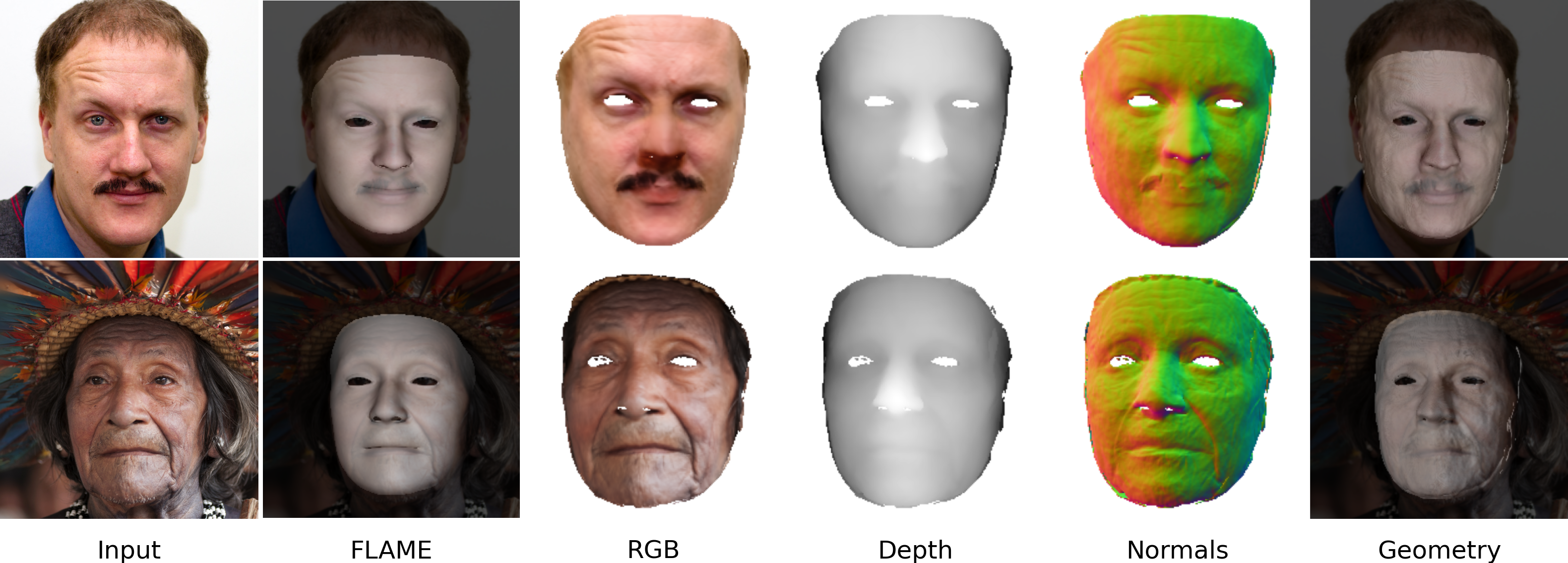}
\end{center}
    \caption{\textbf{Reconstructions by \MethodName:} Above we show the results of \MethodName on images from the FFHQ \cite{StyleGAN} dataset. The first column is the input image with respect to which \MethodName is optimized. The second column is the overlay of coarse FLAME \cite{FLAME:SiggraphAsia2017} mesh on the input image. The third column is the render of the detailed geoemtry learnt by \MethodName. The fourth column is the depth. The fifth column are the normals of the deetailed face geometry learnt by \MethodName and the last column is the overlay of the learnt detailed geometry on top of the input image.}
\label{fig:sider_res}
\end{figure*}

\begin{figure}[h]
\begin{center}
\includegraphics[width=\linewidth]{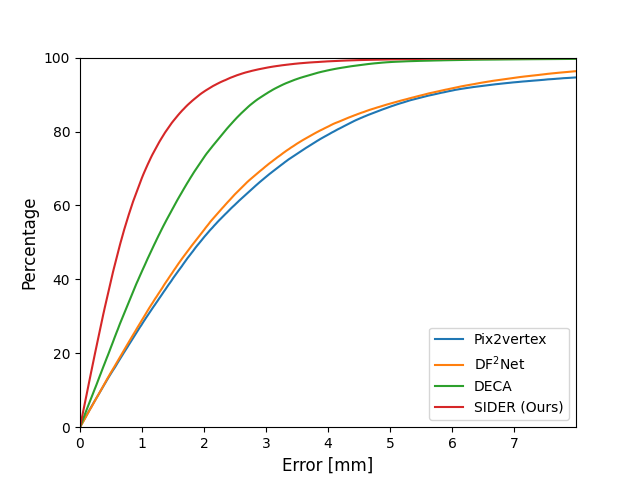}
\end{center}
   \caption{Cumulative error plot. Reconstruction error on the NoW validation set for the different methods.}
   \vspace{-5mm}
\label{fig:quant}
\end{figure}

\begin{figure*}[h]
\begin{center}
\includegraphics[width=\linewidth]{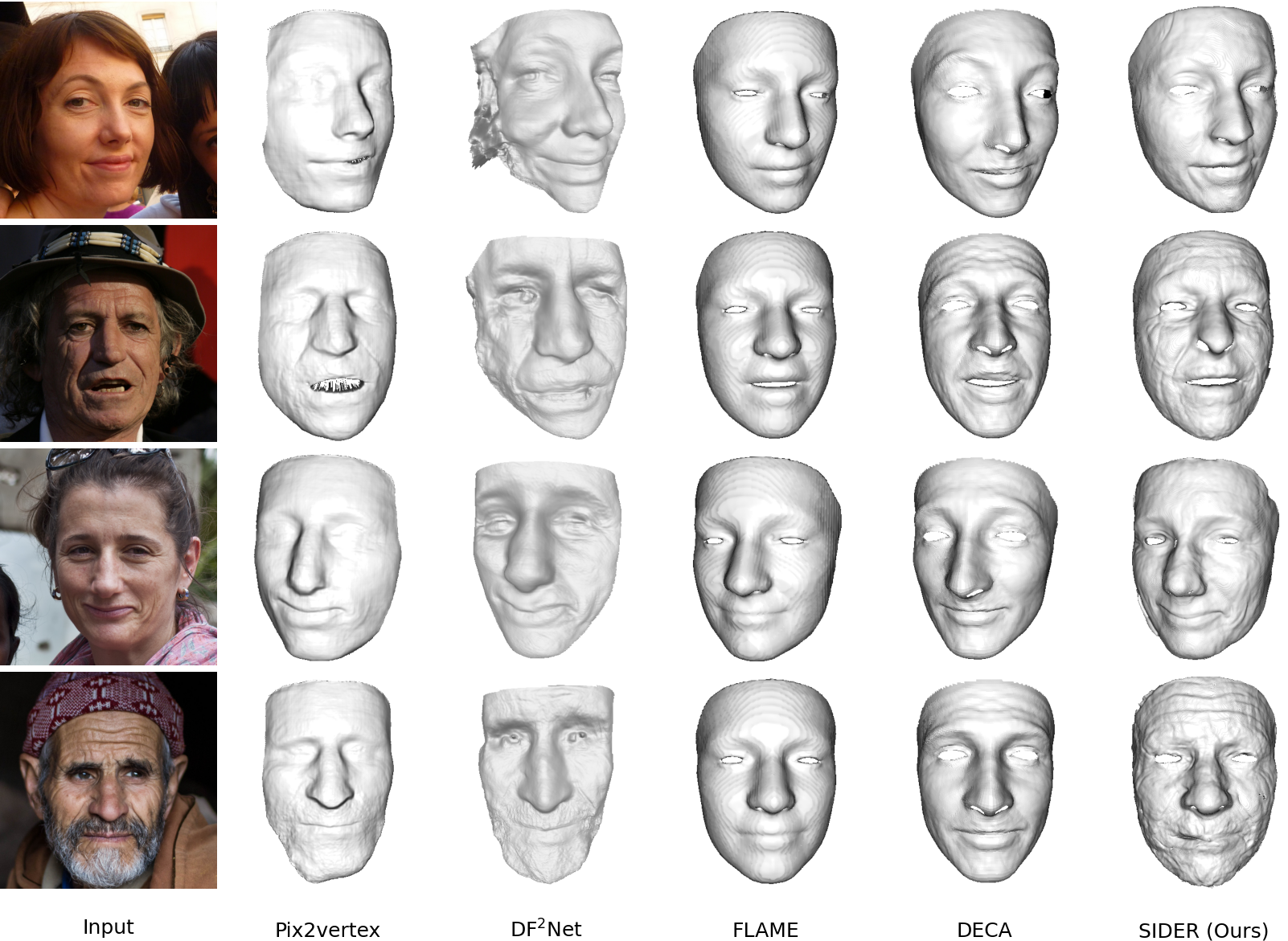}
\end{center}
   \caption{\textbf{Comparison to state-of-the-art methods:} Here we compare again prior art: Pix2vertex~\cite{sela2017unrestricted}, DF$^{2}$Net~\cite{df2net}, FLAME fitting~\cite{FLAME:SiggraphAsia2017} and DECA (w/ details)~\cite{DECA}. Input images taken from FFHQ.}
   \vspace{-5mm}
\label{fig:comp}
\end{figure*}

\begin{figure*}[h]
\begin{center}
\includegraphics[width=\linewidth]{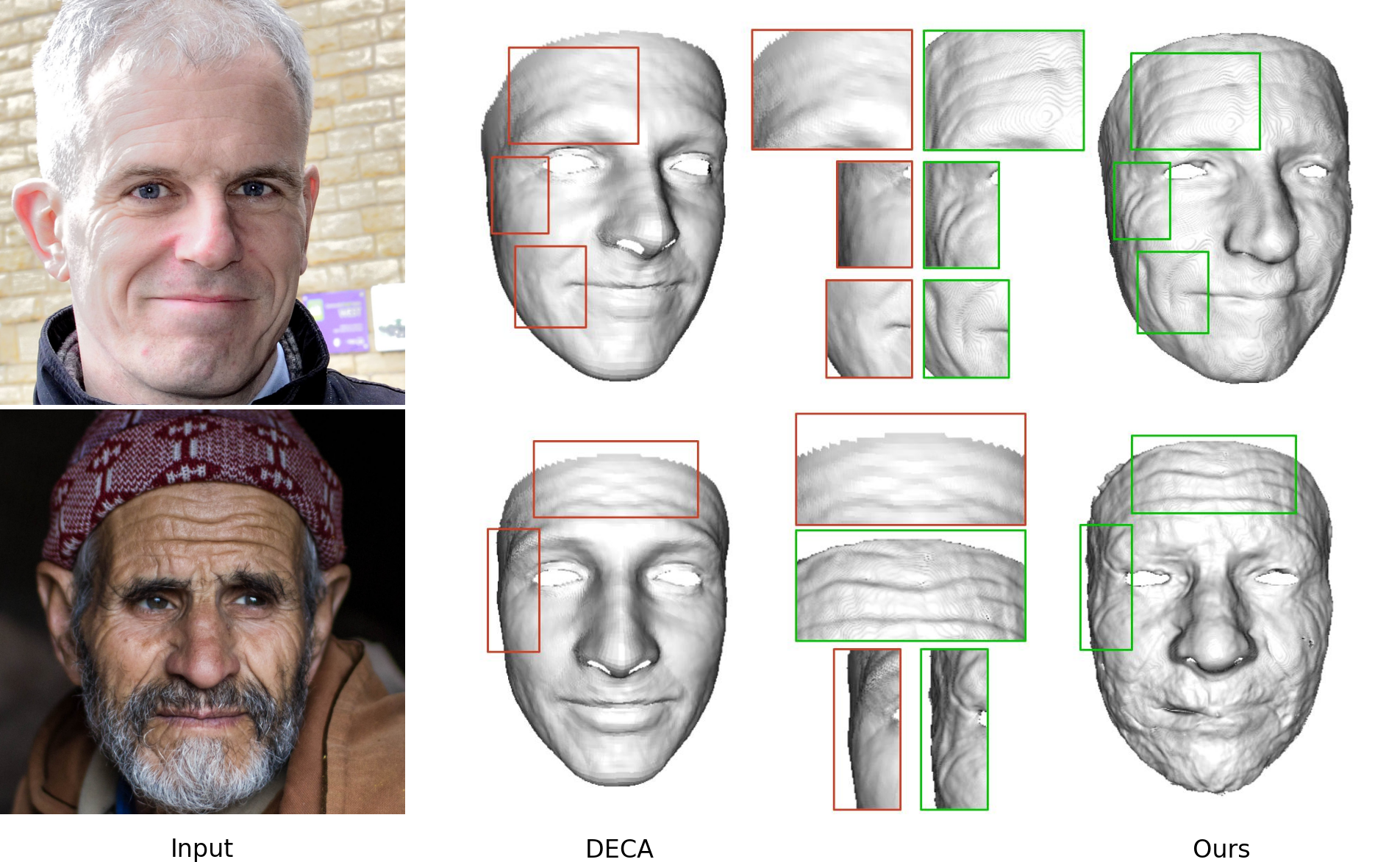}
\end{center}
   \caption{\textbf{Detailed comparison to DECA (w/ details)~\cite{DECA}:} Here we compare the facial geometric details generated by DECA (\textcolor{red}{red} insets) and \MethodName (\textcolor{green}{green} insets). For the image in the top row, we see that \MethodName faithfully recovers the details on the forehead, around the left eye and around the left corner of the lip. In contrast, DECA~\cite{DECA} smoothens the details on the forehead and fails to recover the details around the left eye and the left corner of the lip. Similarly, for the image in row 2, \MethodName is able to accurately recover the details on the forehead and around the left eye, while DECA~\cite{DECA} fails to do so.}
   \vspace{-5mm}
\label{fig:deca_comp_insets}
\end{figure*}

\subsection{Implementation Details}

SIDER learns detailed facial geometry given only a single image. We perform extensive experiments on images from three publicly available datasets, namely FFHQ~\cite{StyleGAN}, ALFW2000~\cite{alfw2000}, and the NoW challenge dataset~\cite{Now}. We choose challenging images that contain both subtle and more complicated facial geometric details, including wrinkles and skin folds. The images are resized to 256x256 resolution before fitting the FLAME model. 

The geometry network, \(f_{\theta}(\cdot)\), consists of 8 linear layers and a skip connection from the input to the fourth layer, similarly to~\cite{IDR}. The rendering network, \(g_{\omega}(\cdot)\), consists of 4 linear layers. Its input is non-linearly mapped to learn high frequencies~\cite{nerf}. Each layer of both MLPs includes 512 hidden units. We first train the geometry network for 1000 epochs to learn the coarse geometry. Then, we fine-tune it via photometric optimization and jointly train the rendering network for around 200-300 epochs, until the loss is not decreasing further.
We use Adam optimizer with a learning rate of $10^{-4}$.

\subsection{Ablation Study}
Gradients from the photometric loss of \eq{color_loss} are back-propagated into the geometry network \(f_{\theta}\) via the feature vector, the normals and the intersection point \(\hat{x}\) (see \cite{IDR}). In this section, we ablate the contribution of the feature vector and the normals by alternatively removing one or the other, and finally both, from the input to the rendering network \(g_{\omega}\). \fig{ablation} shows the results of the ablation study, the first image from the left is the input image on which \MethodName was trained, (a) shows the results when the feature vector is removed from the input to \(g_{\omega}\), (b) shows the results when the normals are removed from the input, (c) shows the results when both are removed, and the last image shows the results of the full model. Without the feature vector as input (see \fig{ablation} (a)) the geometry network does not recover any facial geometric details and the geometry is very smooth. In contrast, without the normals as input (see \fig{ablation} (b)), the geometry network does seem to recover some details but the overall reconstruction suffers significantly. Without either the normals or the feature vector as input (see \fig{ablation} (c)) learning completely fails. Using both the normals and the feature vector (see the last column of \fig{ablation}) as input to the rendering network, \(g_{\omega}\), allows \MethodName to recover high quality facial geometric details without compromising overall reconstruction quality.

\subsection{Quantitative Evaluation}

We evaluate the accuracy of the detailed reconstruction of \MethodName on the NoW challenge dataset~\cite{Now}. We compare it with the performance of recent state-of-the-art methods, namely Pix2vertex~\cite{sela2017unrestricted}, DF$^{2}$Net~\cite{df2net}, and DECA~\cite{DECA}. We choose NoW for our quantitative evaluation, since it includes 3D ground truth scans and provides a standard evaluation protocol. It measures the distance from all reference scan vertices to the closest point in the reconstructed mesh surface, after rigidly aligning scans and reconstructions. Because Pix2vertex~\cite{sela2017unrestricted} and DF$^{2}$Net~\cite{df2net} reconstruct a smaller part of the face than the other methods, we ensure that the ground truth face scan is cropped to a circular area (same for all the methods) that would not exceed the smallest reconstructed mesh.

As shown on~\tab{quant_results}, \MethodName outperforms the current state-of-the-art, i.e. DECA~\cite{DECA}, and other methods by a healthy margin. The mean and median error of \MethodName is 0.89 mm and 0.66 mm respectively; both of which are less than that of DECA~\cite{DECA}, which has a median error 1.19 mm and mean error of 1.47 mm. Additionally, the standard deviation of \MethodName is lower than that of DECA~\cite{DECA}, 0.88 mm as compared to 1.25 mm of DECA~\cite{DECA}. Similarly, the cumulative error plots in~\fig{quant} demonstrate the significant advantage of \MethodName compared to the other methods. 


\begin{table}[t]
\begin{center}
\begin{tabular}{|l|c|c|c|}
\hline
Method & Median (mm) & Mean (mm) 
& Std (mm) \\
\hline\hline
Pix2vertex~\cite{sela2017unrestricted} & 1.93 & 2.74 & 2.85  \\
DF$^{2}$Net~\cite{df2net} & 1.84 & 2.51  & 3.40 \\
DECA~\cite{DECA} & 1.19 & 1.47 & 1.25 \\
\MethodName (Ours) & \textbf{0.66} & \textbf{0.89} & \textbf{0.88} \\
\hline
\end{tabular}
\end{center}
\caption{Quantitative results. Reconstruction error on the NoW validation set.}
\vspace{-7mm}
\label{tab:quant_results}
\end{table}


\subsection{Qualitative Evaluation}
We next provide an extensive qualitative evaluation of SIDER on in-the-wild images  from datasets for which no ground truth is available.

In \fig{sider_res}, we demonstrate qualitative results of \MethodName on images from the FFHQ dataset \cite{StyleGAN}. The columns of the figure show correspondingly: the input image, the FLAME fitting, the learnt render (RGB output), the depth, the normals, and finally the overlay of the reconstructed face, with geometric details, on top of the original image. As can be seen, \MethodName recovers accurately the geometric details for each input image. For example, in the first row of \fig{sider_res}, the geometric details arising on the left side of the forehead due to the raised eyebrow are captured faithfully. Similarly, in the second row we see the wrinkling around the mouth is realistically recovered.

In~\fig{sider_train}, we show the RGB output and normals during the joint training of the geometry and rendering networks. We can see how the networks gradually learn a detailed facial geometry represented as an SDF.

In \fig{comp}, we qualitatively compare the results of \MethodName to state-of-the-art detailed facial reconstruction methods, namely Pix2vertex~\cite{sela2017unrestricted}, DF$^{2}$Net~\cite{df2net}, and DECA~\cite{DECA}. The first column corresponds to the input image, the second column contains the results of Pix2vertex~\cite{sela2017unrestricted}, the third contains the results of DF$^{2}$Net~\cite{df2net}, the fourth column contains the results of simple FLAME fitting~\cite{FLAME:SiggraphAsia2017} (used as ground truth of the coarse geometry during the first stage), the fifth column contains the results of DECA with details~\cite{DECA}, the current state-of-the-art, and the last column contains the results of our method. Note that DECA results are masked with FLAME, in order to illustrate only the face region for fair comparison with the other methods. As can be seen, \MethodName is able to recover significantly more detail than the other methods. Pix2vertex~\cite{sela2017unrestricted} generates detailed reconstructions that are quite smooth and misses large visible geometric details such at the skin fold on the right side of the lip of the image in row 1 or the wrinkles on the forehead of the image in row 4. The results of DF$^{2}$Net~\cite{df2net}, while recovering greater geometric details than Pix2vertex~\cite{sela2017unrestricted}, are still prone to errors. For example, it misses the wrinkles on the forehead of the image in row 2 and recovers incorrect details on the forehead from the image in row 4. The produced meshes also contain artifacts, e.g. on the right part of the head in rows 1 and 2, or include extreme curvature, e.g. in the eyes region. The face shape is also affected by lighting to a great extent (e.g. shaded part in row 2). 

DECA~\cite{DECA}, the current state-of-the-art in facial geometric details recovery, generates better reconstructions than both Pix2vertex~\cite{sela2017unrestricted} and DF$^{2}$Net~\cite{df2net}, however it is unable to recover fine geometric details. For example, the skin fold on the left side of the mouth of the image in row 1 is captured by both our method, \MethodName, and  DF$^{2}$Net~\cite{df2net}. In contrast, DECA~\cite{DECA} is unable to recover it. Similarly, the geometric details under the left eye of the image in row 2 are not recovered by DECA~\cite{DECA}. \MethodName, however, is able to accurately recover them. The skin folds around the eyes and around the lips of the image in row 3 are faithfully recovered by \MethodName, but DECA~\cite{DECA} is unable to reconstruct them. In \fig{deca_comp_insets}, we show a direct comparison of the facial geometric details generated by \MethodName and DECA~\cite{DECA}. Insets in green show details generated by \MethodName and insets in red show details generated by DECA~\cite{DECA}. For the image in row 1, we see that \MethodName is faithfully able to recover the details on the forehead, around the eyes, and around the mouth. In contrast, the details recovered by DECA~\cite{DECA} are over-smoothed and  inaccurate. Similarly, for the image in row 2, we see that the details recovered by \MethodName have greater fidelity to the input than the details recovered by DECA~\cite{DECA}.

In summary, as can be seen from the qualitative results in both \fig{sider_res}, \fig{deca_comp_insets} and \fig{comp}, \MethodName is able to reconstruct high-quality facial geometric details from single images, more accurately and with greater fidelity to the input image than competing methods.

%% file: suppmat_arxiv.tex
\begin{center}
    \section*{\textbf{\MethodName \includegraphics[height=15pt]{images/cider.jpg}: Single-Image Neural Optimization for Facial Geometric Detail Recovery}\\
 - \textbf{Supplementary} -}
\end{center}

\maketitle
\begin{figure*}[ht]
\begin{center}
\includegraphics[width=\linewidth]{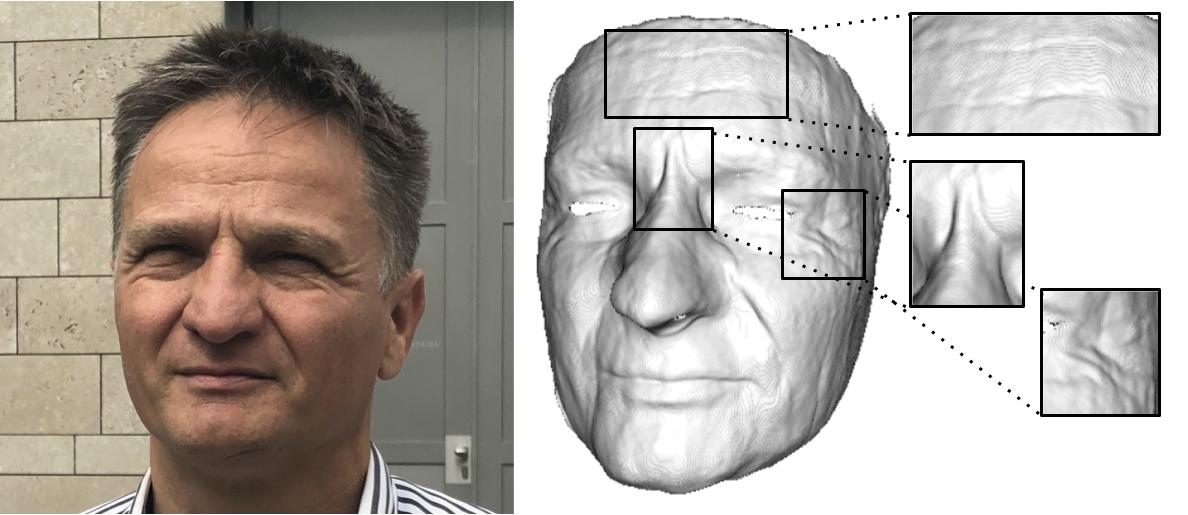}
\end{center}
   \caption{\textbf{Reconstructions by \MethodName:} \textbf{SIDER}  is a novel photometric optimization method that recovers, from a single image, detailed facial geometry without any 3D, multi-view or multiple image supervision. As shown above, the details recovered by SIDER such as wrinkles, skin folds and skin bumps are realistic and have high fidelity to the input image. Please note the enlarged insets of the geometric details on the forehead and the eyes (both on the center and on the right). The image is taken from NoW dataset \cite{Now}}
\label{fig:teas1}
\end{figure*}
\section{More Reconstruction Results}
\begin{figure*}[ht]
\begin{center}
\includegraphics[width=\linewidth]{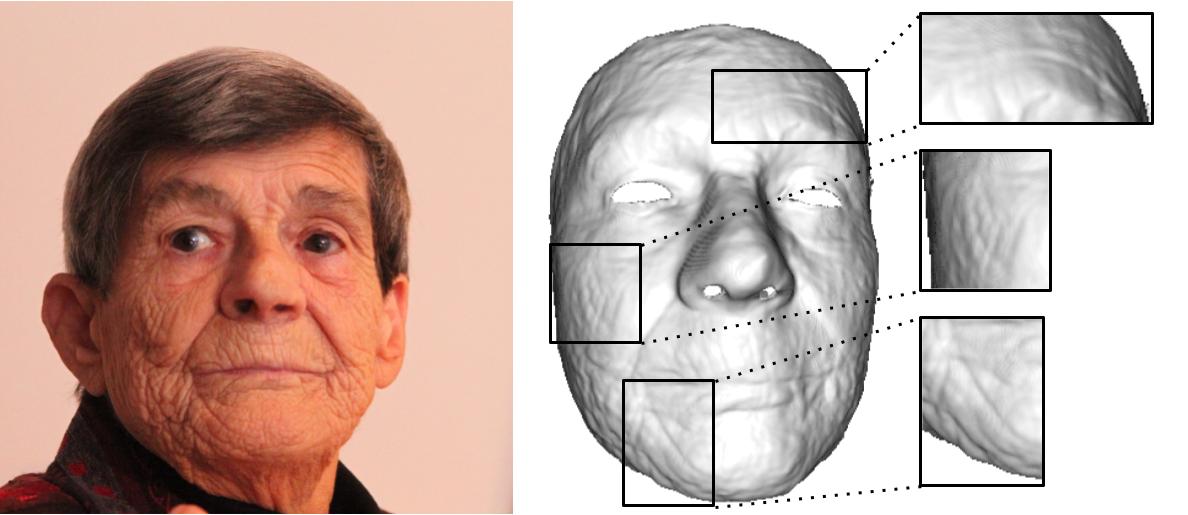}
\includegraphics[width=\linewidth]{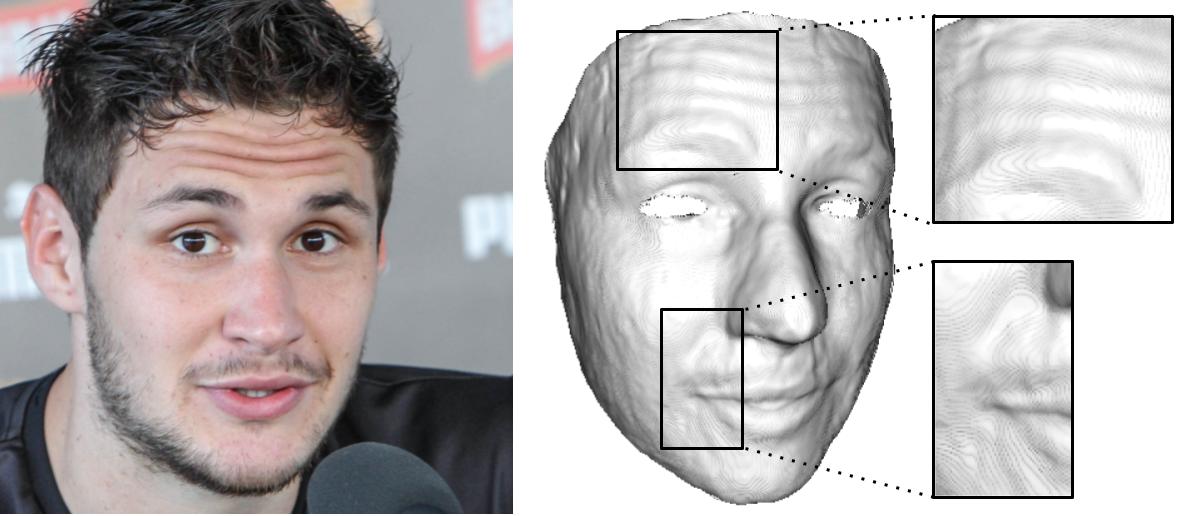}
\includegraphics[width=\linewidth]{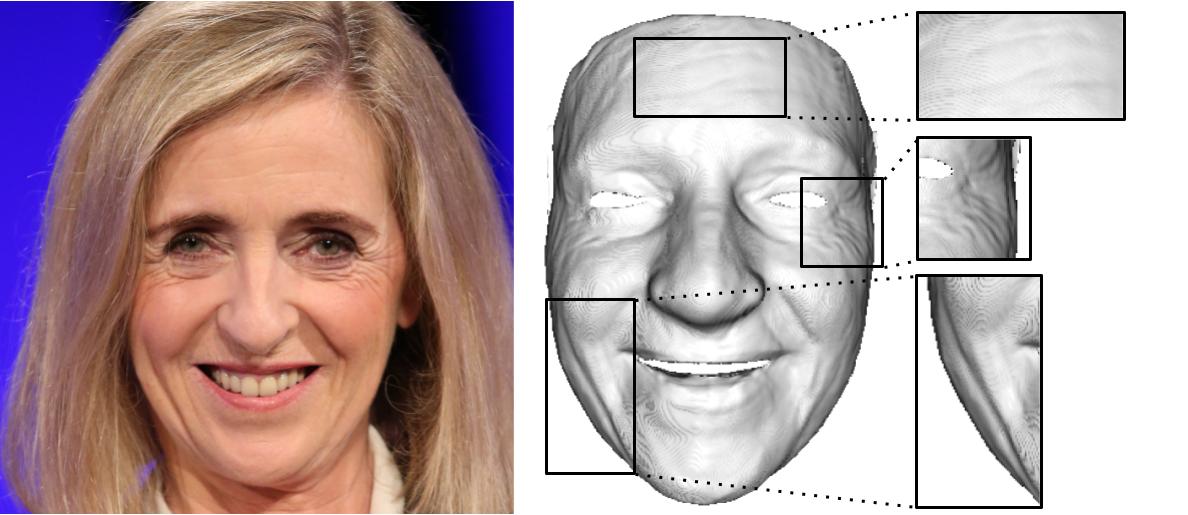}
\end{center}
   \caption{\textbf{Reconstructions by \MethodName:} As shown above, the details recovered by SIDER such as wrinkles, skin folds and skin bumps are realistic and have high fidelity to the input image. Please note the enlarged insets of the geometric details on the forehead and mouth of all three images. The images are taken from the FFHQ dataset \cite{StyleGAN}}
\label{fig:teas2}
\end{figure*}

In this section, we provide more reconstruction results by \MethodName. In \fig{teas1} and \fig{teas2}, we show sample reconstructions of \MethodName from the NoW dataset \cite{Now} and the FFHQ dataset \cite{StyleGAN} respectively. In \fig{teas1}, we see that the geometric details on the forehead and on around the eyes (both on the center and on the right) are accurately recovered. In \fig{teas2}, we see that the geometric details on the forehead and on around the mouth are accurately recovered.

\begin{figure*}[ht]
\begin{center}
\vspace{-2cm}
\includegraphics[width=\linewidth]{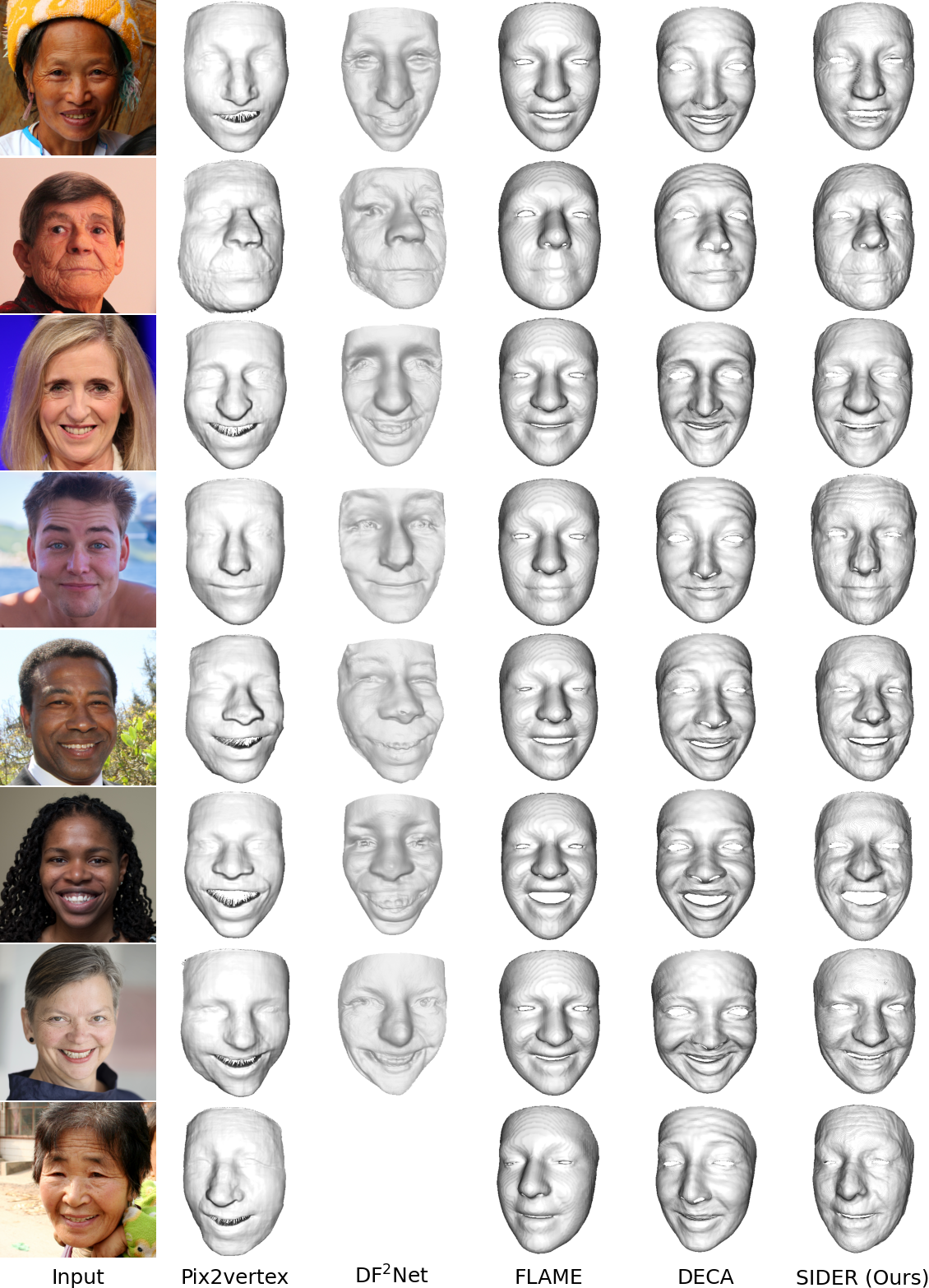}
\end{center}
   \caption{\textbf{Comparison to state-of-the-art methods:} Here we compare \MethodName against prior art. From left to right we see results of Pix2vertex~\cite{sela2017unrestricted}, DF$^{2}$Net~\cite{df2net}, FLAME fitting~\cite{FLAME:SiggraphAsia2017}, DECA (with details)~\cite{DECA} and \MethodName. Blank entries indicate that the particular method did not return any reconstruction. As can be seen, across all images, \MethodName recovers high quality facial geometric details that have a high fidelity to the input image. Note the accurate recovery of facial geometric details around the mouth of the images in the second, third and seventh row.}
   \vspace{-5mm}
\label{fig:sup1}
\end{figure*}

\begin{figure*}[ht]
\begin{center}
\vspace{-2cm}
\includegraphics[width=\linewidth]{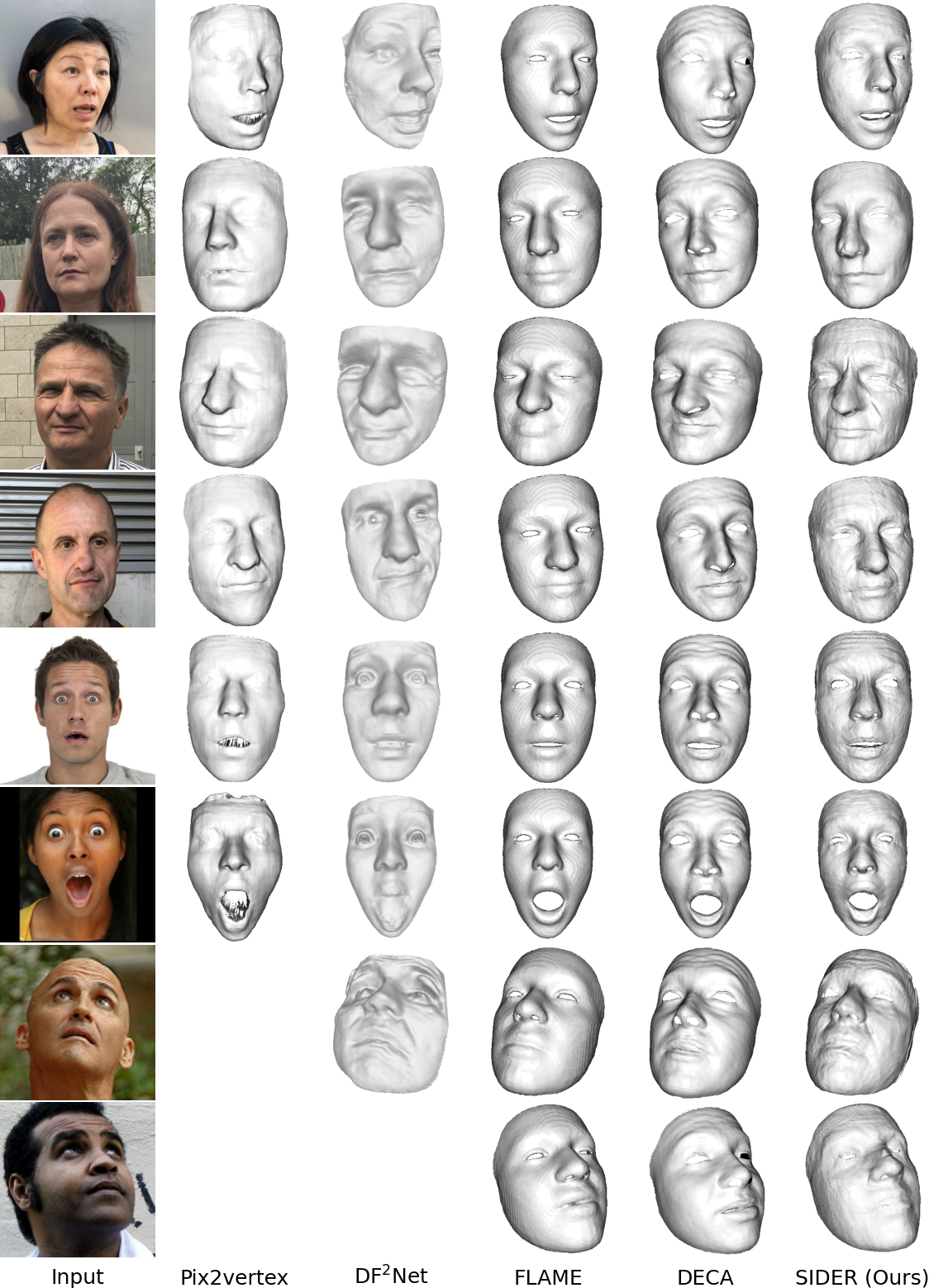}
\end{center}
   \caption{\textbf{Comparison to state-of-the-art methods:} Here we compare \MethodName against prior art. From left to right we see results of Pix2vertex~\cite{sela2017unrestricted}, DF$^{2}$Net~\cite{df2net}, FLAME fitting~\cite{FLAME:SiggraphAsia2017}, DECA (with details)~\cite{DECA} and \MethodName. Blank entries indicate that the particular method did not return any reconstruction. As can be seen, across all images, \MethodName recovers high quality facial geometric details that have a high fidelity to the input image. Note the accurate recovery of facial geometric details on the forehead of the images in the seventh and the eighth row despite the extreme head-pose.}
   \vspace{-5mm}
\label{fig:sup2}
\end{figure*}

In \fig{sup1} and \fig{sup2}, we compare the results of \MethodName with prior art. In the second, third and seventh row of \fig{sup1}, we see that \MethodName is accurately able to recover the geometric details around the mouth. In the fourth and and eighth row of \fig{sup1}, we see that \MethodName is accurately able to recover the geometric details on the forehead. Similarly, in seventh and eighth row of \fig{sup2} we see accurate recovery of facial geometric details by \MethodName around both the forehead and the mouth despite the extreme head-pose. In summary, \MethodName is able to recover realistic facial geometric details with high fidelity to the input.

%% file: main.bbl
\begin{thebibliography}{10}\itemsep=-1pt

\bibitem{blanz1999morphable}
Volker Blanz, Thomas Vetter, et~al.
\newblock A morphable model for the synthesis of 3d faces.
\newblock 1999.

\bibitem{booth20173d}
James Booth, Epameinondas Antonakos, Stylianos Ploumpis, George Trigeorgis,
  Yannis Panagakis, and Stefanos Zafeiriou.
\newblock 3d face morphable models" in-the-wild".
\newblock In {\em CVPR}. IEEE, 2017.

\bibitem{USDFs}
Julian Chibane, A. Mir, and Gerard Pons-Moll.
\newblock Neural unsigned distance fields for implicit function learning.
\newblock In {\em NeurIPS}, 2020.

\bibitem{dou2017end}
Pengfei Dou, Shishir~K Shah, and Ioannis~A Kakadiaris.
\newblock End-to-end 3d face reconstruction with deep neural networks.
\newblock In {\em CVPR}, 2017.

\bibitem{DECA}
Yao Feng, Haiwen Feng, Michael~J. Black, and Timo Bolkart.
\newblock Learning an animatable detailed {3D} face model from in-the-wild
  images.
\newblock volume~40, 2021.

\bibitem{Genova_2018_CVPR}
Kyle Genova, Forrester Cole, Aaron Maschinot, Aaron Sarna, Daniel Vlasic, and
  William~T. Freeman.
\newblock Unsupervised training for 3d morphable model regression.
\newblock In {\em CVPR}, 2018.

\bibitem{KANface}
Markos Georgopoulos, Yannis Panagakis, and Maja Pantic.
\newblock Investigating bias in deep face analysis: The kanface dataset and
  empirical study, 2020.

\bibitem{IGR}
Amos Gropp, Lior Yariv, Niv Haim, Matan Atzmon, and Yaron Lipman.
\newblock Implicit geometric regularization for learning shapes.
\newblock In {\em ICML}, pages 3569--3579. 2020.

\bibitem{3DDFA_V2}
Jianzhu Guo, Xiangyu Zhu, Yang Yang, Fan Yang, Zhen Lei, and Stan~Z Li.
\newblock Towards fast, accurate and stable 3d dense face alignment.
\newblock In {\em Proceedings of the European Conference on Computer Vision
  (ECCV)}, 2020.

\bibitem{jackson2017large}
Aaron~S Jackson, Adrian Bulat, Vasileios Argyriou, and Georgios Tzimiropoulos.
\newblock Large pose 3d face reconstruction from a single image via direct
  volumetric cnn regression.
\newblock In {\em ICCV}, 2017.

\bibitem{StyleGAN}
Tero Karras, Samuli Laine, and Timo Aila.
\newblock A style-based generator architecture for generative adversarial
  networks.
\newblock In {\em CVPR}, 2019.

\bibitem{kim2018inversefacenet}
Hyeongwoo Kim, Michael Zollh{\"o}fer, Ayush Tewari, Justus Thies, Christian
  Richardt, and Christian Theobalt.
\newblock Inversefacenet: Deep monocular inverse face rendering.
\newblock In {\em CVPR}, 2018.

\bibitem{FLAME:SiggraphAsia2017}
Tianye Li, Timo Bolkart, Michael.~J. Black, Hao Li, and Javier Romero.
\newblock Learning a model of facial shape and expression from {4D} scans.
\newblock {\em ACM Transactions on Graphics, (Proc. SIGGRAPH Asia)}, 36(6),
  2017.

\bibitem{Liu2019LearningTI}
Shichen Liu, S. Saito, Weikai Chen, and H. Li.
\newblock Learning to infer implicit surfaces without 3d supervision.
\newblock In {\em NeurIPS}, 2019.

\bibitem{OccNet}
Lars Mescheder, Michael Oechsle, Michael Niemeyer, Sebastian Nowozin, and
  Andreas Geiger.
\newblock Occupancy networks: Learning 3d reconstruction in function space.
\newblock In {\em CVPR}, 2019.

\bibitem{nerf}
Ben Mildenhall, Pratul~P Srinivasan, Matthew Tancik, Jonathan~T Barron, Ravi
  Ramamoorthi, and Ren Ng.
\newblock Nerf: Representing scenes as neural radiance fields for view
  synthesis.
\newblock 2020.

\bibitem{DVR}
Michael Niemeyer, Lars Mescheder, Michael Oechsle, and Andreas Geiger.
\newblock Differentiable volumetric rendering: Learning implicit 3d
  representations without 3d supervision.
\newblock In {\em CVPR}, 2020.

\bibitem{DeepSDF}
Jeong~Joon Park, Peter Florence, Julian Straub, Richard Newcombe, and Steven
  Lovegrove.
\newblock Deepsdf: Learning continuous signed distance functions for shape
  representation.
\newblock In {\em CVPR}, June 2019.

\bibitem{Richardson2017FaceRF}
Elad Richardson, Matan Sela, Roy Or-El, and R. Kimmel.
\newblock Face reconstruction from a single image.
\newblock In {\em CVPR}, 2017.

\bibitem{Now}
Soubhik Sanyal, Timo Bolkart, Haiwen Feng, and Michael Black.
\newblock Learning to regress 3d face shape and expression from an image
  without 3d supervision.
\newblock In {\em Proceedings IEEE Conf. on Computer Vision and Pattern
  Recognition (CVPR)}, June 2019.

\bibitem{sela2017unrestricted}
Matan Sela, Elad Richardson, and Ron Kimmel.
\newblock Unrestricted facial geometry reconstruction using image-to-image
  translation.
\newblock In {\em ICCV}, 2017.

\bibitem{siren}
Vincent Sitzmann, Julien~N.P. Martel, Alexander~W. Bergman, David~B. Lindell,
  and Gordon Wetzstein.
\newblock Implicit neural representations with periodic activation functions.
\newblock In {\em NeurIPS}, 2020.

\bibitem{nglod}
Towaki Takikawa, Joey Litalien, Kangxue Yin, Karsten Kreis, Charles Loop, Derek
  Nowrouzezahrai, Alec Jacobson, Morgan McGuire, and Sanja Fidler.
\newblock Neural geometric level of detail: Real-time rendering with implicit
  {3D} shapes.
\newblock 2021.

\bibitem{tewari2019fml}
Ayush Tewari, Florian Bernard, Pablo Garrido, Gaurav Bharaj, Mohamed Elgharib,
  Hans-Peter Seidel, Patrick P{\'e}rez, Michael Z{\"o}llhofer, and Christian
  Theobalt.
\newblock Fml: Face model learning from videos.
\newblock In {\em Proceedings of the IEEE Conference on Computer Vision and
  Pattern Recognition}, pages 10812--10822, 2019.

\bibitem{tewari2017self}
Ayush Tewari, Michael Zollh{\"o}fer, Pablo Garrido, Florian Bernard, Hyeongwoo
  Kim, Patrick P{\'e}rez, and Christian Theobalt.
\newblock Self-supervised multi-level face model learning for monocular
  reconstruction at over 250 hz.
\newblock In {\em CVPR}, 2018.

\bibitem{tran2019towards}
Luan Tran, Feng Liu, and Xiaoming Liu.
\newblock Towards high-fidelity nonlinear 3d face morphable model.
\newblock In {\em In Proceeding of IEEE Computer Vision and Pattern
  Recognition}, Long Beach, CA, June 2019.

\bibitem{tran2018nonlinear}
Luan Tran and Xiaoming Liu.
\newblock Nonlinear 3d face morphable model.
\newblock In {\em IEEE Computer Vision and Pattern Recognition (CVPR)}, Salt
  Lake City, UT, June 2018.

\bibitem{extreme3D}
Anh Tuấn~Trần, Tal Hassner, Iacopo Masi, Eran Paz, Yuval Nirkin, and
  G{\'e}rard Medioni.
\newblock Extreme 3d face reconstruction: Seeing through occlusions.
\newblock In {\em CVPR}, 2018.

\bibitem{IDR}
Lior Yariv, Yoni Kasten, Dror Moran, Meirav Galun, Matan Atzmon, Basri Ronen,
  and Yaron Lipman.
\newblock Multiview neural surface reconstruction by disentangling geometry and
  appearance.
\newblock {\em Advances in Neural Information Processing Systems}, 33, 2020.

\bibitem{df2net}
Xiaoxing Zeng, Xiaojiang Peng, and Yu Qiao.
\newblock Df2net: A dense-fine-finer network for detailed 3d face
  reconstruction.
\newblock In {\em Proceedings of the IEEE International Conference on Computer
  Vision}, pages 2315--2324, 2019.

\bibitem{reda}
Wenbin Zhu, HsiangTao Wu, Zeyu Chen, Noranart Vesdapunt, and Baoyuan Wang.
\newblock Reda:reinforced differentiable attribute for 3d face reconstruction.
\newblock In {\em Proceedings of the IEEE/CVF Conference on Computer Vision and
  Pattern Recognition (CVPR)}, June 2020.

\bibitem{alfw2000}
Xiangyu Zhu, Zhen Lei, Junjie Yan, Dong Yi, and Stan~Z. Li.
\newblock High-fidelity pose and expression normalization for face recognition
  in the wild.
\newblock In {\em 2015 IEEE Conference on Computer Vision and Pattern
  Recognition (CVPR)}, pages 787--796, 2015.

\end{thebibliography}
